\documentclass[letterpaper]{article} 
\usepackage{aaai2026}  
\usepackage{times}  
\usepackage{helvet}  
\usepackage{courier}  
\usepackage[hyphens]{url}  
\usepackage{graphicx} 
\usepackage{natbib}  
\usepackage{caption} 
\usepackage{algorithm}
\usepackage{algorithmic}

\usepackage{amsfonts}
\usepackage{amsmath}

\usepackage{newfloat}
\usepackage{listings}
\DeclareCaptionStyle{ruled}{labelfont=normalfont,labelsep=colon,strut=off} 
\floatstyle{ruled}
\newfloat{listing}{tb}{lst}{}
\floatname{listing}{Listing}
\title{StreamingTalker: Audio-driven 3D Facial Animation \\ with Autoregressive Diffusion Model}
\author{
    Yifan Yang\textsuperscript{\rm 1, \rm2},
    Zhi Cen\textsuperscript{\rm 1},
    Sida Peng\textsuperscript{\rm 1},
    Xiangwei Chen\textsuperscript{\rm 3},\\
    Yifu Deng\textsuperscript{\rm 2},
    Xinyu Zhu\textsuperscript{\rm 2},
    Fan Jia\textsuperscript{\rm 2},
    Xiaowei Zhou\textsuperscript{\rm 1},
    Hujun Bao\textsuperscript{\rm 1}\thanks{Corresponding author.}
}
\affiliations{
    \textsuperscript{\rm 1}State Key Laboratory of CAD\&CG, Zhejiang University\\
    \textsuperscript{\rm 2}Ant Group\\
    \textsuperscript{\rm 3}College of Computer Science, Zhejiang University\\

    yfyang@zju.edu.cn
}

\usepackage{bibentry}

\usepackage{xcolor}
\usepackage{booktabs} 
\usepackage{multirow} 

\newcommand{\PAR}[1]{\vskip4pt \noindent{\bf #1~}}

\def\Tabref#1{Table~\ref{#1}}
\def\Figref#1{Figure~\ref{#1}}
\def\Secref#1{Section~\ref{#1}}
\def\Eqref#1{Equation~\ref{#1}}

\newcommand{\seqlen}{T}
\newcommand{\vertdim}{V}
\newcommand{\nembed}{N}
\newcommand{\channel}{C}
\newcommand{\fcompo}{H}

\newcommand{\xbold}{\mathbf{x}}

\newcommand{\abold}{\mathbf{a}}

\newcommand{\style}{s_k}

\newcommand{\zprm}{\mathbf{z}}

\newcommand{\encoder}{E}
\newcommand{\decoder}{D}

\newcommand{\zq}{\zprm_q}
\newcommand{\zhat}{\hat{\zprm}}
\newcommand{\codebook}{\mathcal{Z}}
\newcommand{\quantize}{quantize}

\newcommand{\windowsize}{h}

\newcommand{\timestep}{t}
\newcommand{\noisestep}{N}

\newcommand{\durationl}{\seqlen-\windowsize+1:\seqlen}
\newcommand{\duration}{\seqlen-\windowsize:\seqlen-1}
\newcommand{\duraplus}{\seqlen-\windowsize+1:\seqlen}
\newcommand{\duraminus}{\seqlen-\windowsize:\seqlen-1}

\newcommand{\condition}{C_{A}}
\newcommand{\sample}[2]{q(#1|#2)}

\begin{document}

\maketitle


\begin{abstract}
This paper focuses on the task of speech-driven 3D facial animation, which aims to generate realistic and synchronized facial motions driven by speech inputs.
Recent methods have employed audio-conditioned diffusion models for 3D facial animation, achieving impressive results in generating expressive and natural animations.
However, these methods process the whole audio sequences in a single pass, which poses two major challenges: 
they tend to perform poorly when handling audio sequences that exceed the training horizon and will suffer from significant latency when processing long audio inputs.
To address these limitations, we propose a novel autoregressive diffusion model that outputs facial motions in a streaming manner. 
This design ensures flexibility with varying audio lengths and achieves low latency independent of audio duration.
Specifically, we select a limited number of past frames as historical motion context and combine them with the audio input to create a dynamic condition. 
This condition guides a lightweight diffusion head to iteratively generate facial motion frames, enabling real-time synthesis with high-quality results.
Experiments conducted on public datasets demonstrate that our approach outperforms recent baseline methods.
We will release the code at \textcolor{magenta}{\texttt{https://zju3dv.github.io/StreamingTalker/}}.
\end{abstract}    
\section{Introduction}



Speech-driven 3D facial animation has emerged as a critical area of research, garnering significant attention due to its wide-ranging applications in virtual reality, gaming, and telecommunication. 
This technology enables lifelike digital humans that can interact naturally with users through speech, enhancing both immersion and accessibility.
The goal of this work is to generate a 3D facial mesh that accurately reflects the given speech audio, ensuring precise lip-synchronization and expressive motion.
However, this task remains challenging, as it demands not only high visual fidelity in facial expressions but also efficient processing to support real-time generation.

Traditional approaches \cite{massaro1212,edwards2016jali} to speech-driven 3D facial animation have predominantly employed deterministic methods that directly map phonemes to the corresponding 3D facial mesh.
However, these methods often rely heavily on intermediate representations of phonemes and primarily focus on modeling mouth movements, neglecting other facial dynamics.
Moreover, these traditional methods lack the capability for one-to-many generations, which significantly limits their ability to capture the variability and expressiveness of natural facial motion.

In recent years, diffusion models have achieved remarkable success in image generation tasks \cite{ramesh2021zero}, leading to their introduction in the domain of speech-driven facial animation \cite{ma2024diffspeaker,sun2024diffposetalk}.
Notable methods such as FaceDiffuser \cite{stan2023facediffuser} have demonstrated the potential of diffusion processes in this context, achieving fast denoising and good results on short sequences.
However, we observe that these designs exhibit a performance drop when generating motions that extend beyond the training horizon. 
This limitation arises because they are trained on fixed-length sequences, which hinders their ability to generalize to longer and more complex sequences.
Furthermore, as these methods employ large denoising networks such as Transformers or U-Nets and need to process the entire sequence before producing results, they incur heavy computation and struggle to support real-time applications.

Several critical obstacles must be addressed to enable real-time generation of arbitrary-length facial meshes.
First, it is essential to develop a mechanism that dynamically integrates historical information from past motions, allowing the model to handle streaming outputs effectively.
Additionally, maintaining a balance between generation quality and inference speed remains a key difficulty.

To tackle these challenges, we propose a novel approach that leverages an autoregressive (AR) diffusion model.
Specifically, our method reformulates full-sequence generation as an AR diffusion process conditioned on historical motion.
By encoding a limited number of past frames and the current audio input with an AR transformer, we derive a dynamic condition that guides a lightweight MLP-based diffusion head to generate facial meshes in real time.
This approach offers several key advantages, including the ability to enable the flexible and dynamic use of past motions which enhances the model's generalization capabilities, and the facilitation of real time generation of facial meshes for arbitrary-length sequences, supporting streaming output and ensuring seamless rendering.

Experimental results demonstrate that our approach outperforms state-of-the-art methods on two benchmark datasets, establishing its effectiveness and robustness.
Furthermore, our method achieves superior performance in long-sequence generation and greatly improves inference latency compared with previous diffusion-based models.
\section{Related Work}
\subsection{Speech-driven 3D Facial Animation}\label{sec:animation}

Existing approaches to speech-driven 3D facial animation can be broadly categorized into linguistics-based and learning-based methods. 
Linguistics-based methods typically establish a comprehensive rules to govern the animation process. For example, \cite{massaro1212} uses dominance functions to map phonemes to facial movements, while JALI \cite{edwards2016jali} factors mouth movements into lip and jaw animation.
However, these methods still rely on intermediate representations of phonemes and primarily focus on mouth movements. 
In contrast, learning-based methods have emerged to address these limitations. 
For instance, \cite{taylor2017deep} employs deep neural networks to transform phoneme transcriptions into facial animation parameters, proposing a sliding window approach instead of an RNN. Another notable study \cite{karras2017audio} utilizes convolutional neural networks to animate faces directly from audio data.

Recent studies related to our work have concentrated on training neural networks using cross-modal datasets that combine audio and 3D facial meshes. 
VOCA \cite{cudeiro2019voca}, for example, inputs raw audio and speaker style, represented by subject identity, and utilizes temporal convolutions to animate a static mesh template. 
MeshTalk \cite{richard2021meshtalk} creates a categorical latent space to distinguish between audio-correlated and uncorrelated facial movements. 
Both tend to overlook long-term audio context due to their reliance on short audio windows. 
FaceFormer \cite{fan2022faceformer} addresses this issue by considering long-term audio context with a transformer decoder \cite{vaswani2017attention}, enabling it to generate temporally stable animations. In addition, it employs Wav2Vec2.0 \cite{baevski2020wav2vec}, a self-supervised pre-trained speech model, which helps mitigate the scarcity of data in existing audio-visual datasets by leveraging large-scale unlabeled speech data to learn rich acoustic and linguistic features.  
CodeTalker \cite{xing2023codetalker} integrates a temporal autoregressive model with a latent codebook using VQ-VAE \cite{van2017vqvae}, inspired by Learning2Listen \cite{ng2022learning2listen}. The above-mentioned methods are deterministic models so their diversity is limited since human speech and facial expressions are variable and dynamic.

\subsection{Diffusion Models for Facial Motion Synthesis}\label{sec:motion}
\begin{figure*}[t]
    \centering
    \includegraphics[trim=0cm 4cm 0cm 0cm,width=\linewidth]{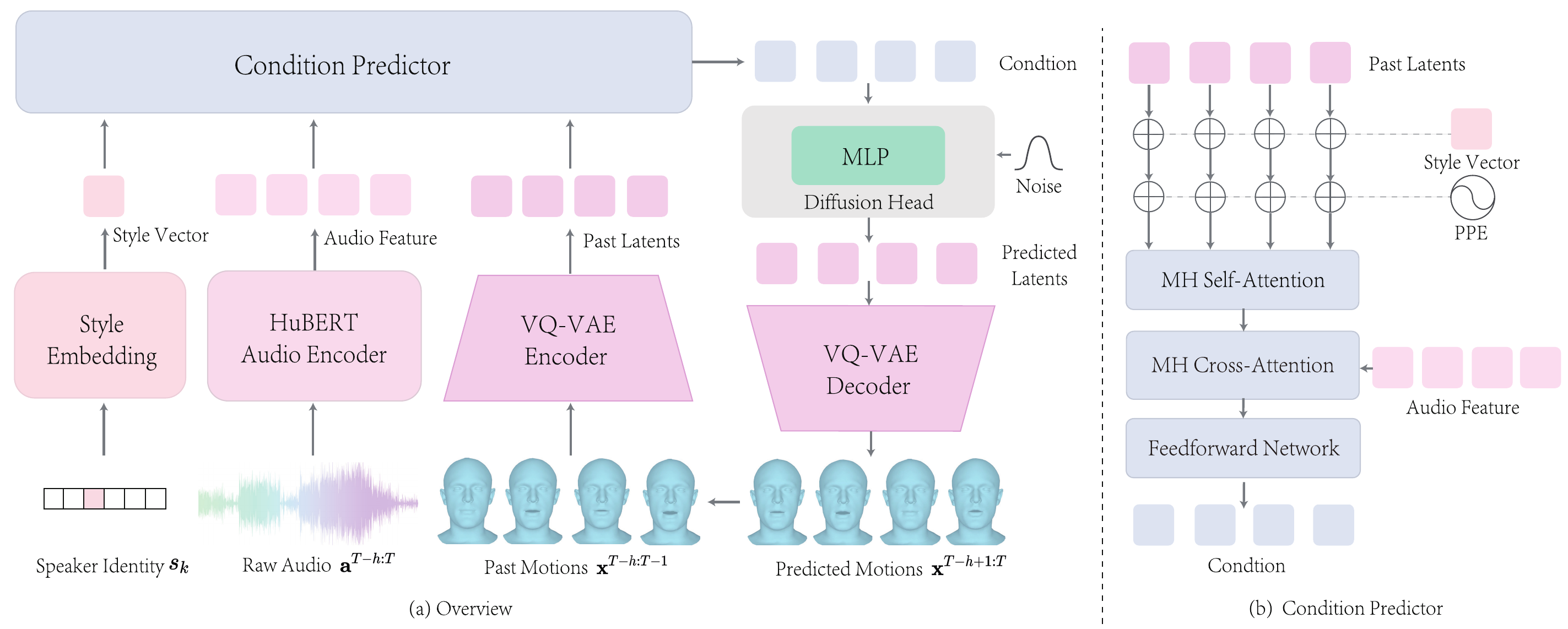}
    \vspace{+25pt}
    \caption{\textbf{(a) Overview of the pipeline.} We employ an AR diffusion model to generate speech-driven 3D facial animations for inputs of arbitrary length. The model first encodes past motions $\xbold^{\duration}$, raw audio $\abold^{\seqlen-\windowsize:\seqlen}$ and speaker identity $\style$ to a dynamic condition. Then the diffusion head leverages this condition to guide the diffusion process.  \textbf{(b) The condition predictor.} The condition predictor uses a transformer network to fuse the motion and audio modalities.}
    \label{fig:pipeline}
\end{figure*}

Diffusion probabilistic models \cite{ho2020denoising,sohl2015diffusion}, which differ from previous generative approaches such as GANs \cite{goodfellow2020gan} and VAEs \cite{kingma2013vae}, have achieved remarkable results in various generative tasks. 
These models adopt a Markov chain to gradually add noise to data samples and subsequently use a neural network to approximate the reverse process to denoise the samples. Given their strong abilities in modeling continuous distributions and one-to-many mapping relationships, diffusion models are particularly suitable for animation tasks.

Diffusion models have been applied in related tasks such as 2D talking face generation \cite{du2023dae}. 
FaceDiffuser \cite{stan2023facediffuser} was the first to introduce diffusion models to the domain of speech-driven 3D facial animation. 
DiffPoseTalk \cite{sun2024diffposetalk} introduces a speaking style encoder and overcomes the limitations of existing diffusion models that cannot be directly transferred to speech-driven expression animation. 
GLDiTalker \cite{lin2024glditalker} proposes a graph-enhanced quantized space and applies a two-stage training strategy. 
DiffSpeaker \cite{ma2024diffspeaker} is one of the most recent and effective works in this field, which introduces a novel biased conditional attention mechanism that utilizes encodings to integrate speaking style and diffusion step information.
However, these works typically denoise the entire sequence together, which can lead to performance drops when generating samples that extend beyond the training horizon. 
In contrast, our model innovates using an autoregressive approach to extract conditions from past motions to guide the diffusion process, enabling flexible generation on arbitrary long sequences.

\section{Method}

Given a speech snippet $\abold ^{1:\seqlen} = (\abold _1,\dots,\abold _\seqlen)$ and a speaker identity  $\style$, our goal is to generate a facial mesh sequence $\xbold ^{1:\seqlen} = (\xbold _1,\dots,\xbold _\seqlen)$ where each frame $\xbold_t\in \mathbb{R}^{V\times 3}$ denotes 3D movements over a template face mesh comprising $\vertdim$ vertices. For simplicity, in the following part, we will denote facial motion by $\xbold$ and audio by $\abold$.

We formulate speech-driven facial animation as a conditional generation problem and propose to use an autoregressive (AR) diffusion model to solve it. An overview of our proposed method is illustrated in \Figref{fig:pipeline}. We begin by converting facial motion to latent space using VQ-VAE in \Secref{sec:latentspace}. To overcome the challenge of integrating past motions, we introduce the AR diffusion model, which use the output of the AR condition predictor to guide the diffusion process in \Secref{sec:ardiffusion}. In \Secref{sec:loss}, we present the training objectives.

\subsection{Learning Latent Space of Facial Motions}\label{sec:latentspace}

To train generative models more easily, it is common to learn a latent space representation of the raw data~\cite{rombach2022}.
CodeTalker \cite{xing2023codetalker} has shown that VQ-VAE can learn a compact and discrete latent space for facial motions.
Inspired by this, we adopt a similar VQ-VAE architecture to encode raw facial motion sequences into latent representations.

Our transformer-based VQ-VAE consists of an encoder $\encoder$, a decoder $\decoder$, and a codebook $\codebook = \left \{ \mathbf{z}_j\in \mathbb{R}^{C} \right \}^{N}_{j=1}$ containing facial motion primitives $\zprm$, where $\nembed$ represents the size of the codebook.
The facial motion data $\xbold$ are initially encoded into a continuous feature vector $\zhat = \encoder(\xbold) \in \mathbb{R}^{\seqlen \times \channel}$. 
This feature vector is then reshaped into $\zhat_h \in \mathbb{R}^{\seqlen' \times \fcompo \times \channel}$ where $\fcompo$ denotes the number of facial components, and $\seqlen'$ is the number of encoded units, given by $\seqlen' = \frac{\seqlen}{\fcompo}$.
The quantized feature vector $\zq \in \mathbb{R}^{\seqlen' \times \fcompo \times \channel}$ is obtained through the quantization operator $\text{\quantize}(\cdot)$, which calculates the distances between each element of $\zhat$ and the entries in the codebook, selecting the closest match as follows:
\begin{equation}
  \zq = \text{\quantize}(\zhat) = \underset{\zprm_k \in \codebook}{\operatorname{arg~min}} \|\zhat_i - \zprm_k\|_2.
  \label{eq:quant}
\end{equation}
Finally, the decoder is employed to reconstruct the facial motions, as defined by \Eqref{eq:selfrecon}:
\begin{equation}
  \hat{\xbold} = D(\zq) = D(\text{quantize}(E(\xbold))).
  \label{eq:selfrecon}
\end{equation}

\subsection{Autoregressive Diffusion Model for Facial Animation}\label{sec:ardiffusion}
 
In the following sections, we detail how our model generates facial meshes based on a speech snippet $\abold$, style $\style$, and time step $\timestep$. 
The process begins with input encoding, 
after which the AR condition predictor estimates the condition for the next frame.
Finally, the diffusion head leverages this condition to guide the denoising process.   
Figure~\ref{fig:pipeline} illustrates the entire process.

We first introduce how we encode inputs. Relative studies \cite{fan2022faceformer,stan2023facediffuser,xing2023codetalker} have shown that self-supervised pre-trained speech model features like
Wav2Vec2 \cite{baevski2020wav2vec} and HuBERT \cite{hsu2021hubert} outperform traditional ones such as MFCC. HuBERT was trained on 960 hours of LibriSpeech \cite{panayotov2015librispeech} dataset and \cite{ma2024diffspeaker} using HuBERT achieves previous state-of-the-art performance. Therefore, we choose HuBERT as the audio encoder in this paper and employ a pre-trained \textit{hubert-base-ls960} version of it. 
The audio encoder takes the audio snippet $\abold$ as input and outputs a sequence of audio embeddings $E_a(\abold) \in \mathbb{R}^{\seqlen \times \channel}$. 
In addition, a simple linear projection layer is used to embed the one-hot style vector $\style$ into the latent space as $E_s(\style) \in \mathbb{R}^{1 \times \channel}$.

\PAR{AR Condition Predictor.} 
Previous work that applies diffusion models typically denoises the entire sequence. 
However, this approach results in a significant performance drop when generating long sequences that extend beyond the training horizon, especially for long sequences of several thousand frames. 
To address the challenge of dynamically integrating historical information from past latents and enhancing the model's generalization capabilities, we propose a novel autoregressive condition predictor. 
This condition will be used later to guide the diffusion generation process.

During both training and inference, we propose a fixed history length strategy, selecting the start frame and take the next $\windowsize$ frames as the past motion $\xbold^{\duration}$, which is encoded into latents $\zprm_{past}$.
Notably, $\windowsize$ is chosen based on dataset characteristics, ranging from 60 to 120 frames.
When the past motion length is less than $\windowsize$ at the beginning of inference, we use the entire sequence:
\begin{equation}
  \zprm_{past} = \text{quantize}(E(\xbold^{\duration})) = \zq^{\duration}.
  \label{eq:ts}
\end{equation}

Then we adopt a transformer decoder to predict the condition $\condition$ for the next frame $\seqlen$ with a teacher forcing scheme:
\begin{equation}
  \condition = \text{TransformerDecoder}(\zprm_{past}, E_a(\abold), E_s(\style)).
  \label{eq:ardecoder}
\end{equation}
where $z_{past}$ is the past motion latents, $E_a(\abold)$ is the audio embeddings, and $E_s(\style)$ is the style embeddings.

As shown in \Figref{fig:pipeline} (b), the past motion latents $\zprm_{past}$ is first passed through a biased causal multi-head (MH) self-attention layer based on ALiBi \cite{press2021train}, with a causal mask ensuring that only current or past information is accessible to prevent information leakage. 
Then, a multi-head cross-attention layer aligns the motion and audio modalities by combining the outputs of the HuBERT encoder and the MH self-attention layer.
A notable design is an alignment mask of the decoder \cite{ma2024diffspeaker}, which ensures proper alignment of the speech and motion modalities.

\PAR{Diffusion Head.} 
We leverage the condition $\condition$ from the AR condition predictor to guide the diffusion process in facial animation generation, which is designed to recover latent representations from Gaussian noise based on a conditional distribution $\sample{\zprm_{\timestep-1}}{\zprm_{\timestep},\condition}$.
Since this distribution depends on the entire dataset and is intractable, we approximate it using a learnable neural network.

In practice, we train a single-layer MLP to approximate this distribution, ensuring real-time generation speed while maintaining accuracy (\Eqref{eq:diffhead}).
Particularly, our model directly predicts the clean sample $\zprm_0$, following MDM~\cite{tevet2209mdm} and EDGE~\cite{tseng2023edge}, as it enables us to offer more precise constraints on facial motions. The denoising process is described as follows:
\begin{equation}
  \tilde{\zprm}_0 = \text{MLP}(\zprm_{\timestep},\condition,E_t(\timestep)),
  \label{eq:diffhead}
\end{equation} where $\zprm_{\timestep}$ is current noisy latent and $E_t(\timestep)$ is the time embedding of $\timestep$, providing the model with temporal information.
After generating the latent $\tilde{\zprm}_0$, we convert it into a facial mesh using the VAE decoder:
\begin{equation}
  \tilde{\xbold} = D(\tilde{\zprm}_0).
  \label{eq:diff2mesh}
\end{equation}

\subsection{Training Loss and Implementation Details}\label{sec:loss}

The training process is divided into two stages: (1) pre-training the VQ-VAE model and (2) training the autoregressive diffusion model. 

\PAR{Stage 1.} We pre-train the VQ-VAE model following the approach in \cite{zhang2018mode,razavi2019vqvae2} for 400 iterations. 
The encoder and decoder of the VQ-VAE are both a six-layer transformer architecture with eight attention heads, 
with its feature dimension set to 1024 and the feedforward network dimension set to 1536.

In this stage, we supervise the training with a motion reconstruction loss and a quantization loss similar to CodeTalker \cite{xing2023codetalker}:
\begin{equation}
  \mathcal{L}_{stage1} = \lambda_{rec}\mathcal{L}_{rec} + \lambda_{quant}\mathcal{L}_{quant},
  \label{eq:stage1loss}
\end{equation} where $\lambda_{rec} = \lambda_{quant} = 1.0$.

The motion reconstruction loss calculates the L1 distance between the reconstructed facial motion sequence $\hat{\xbold}$ and the ground truth facial motion sequence $\xbold$:
\begin{equation}
  \mathcal{L}_{rec} = \| \hat{\xbold} - \xbold \|_1.
  \label{eq:reconloss}
\end{equation}
The quantization loss contains two intermediate code-level losses that reduce the distance between codebook $\codebook$ and embedded features $\zhat$:
\begin{equation}
  \mathcal{L}_{quant} = \| sg(\hat{\zprm}) - \zq \|_2^2 + \beta\| \hat{\zprm} - sg(\zq) \|_2^2,
  \label{eq:quantloss}
\end{equation} where $\beta = 0.25$ is a weighting hyperparameter controlling the update speed of codebook and encoder, and $sg(\cdot )$ stands for a stop-gradient operation which is defined as identity with zero partial derivatives in the backward propagation. Note that the quantization process (\Eqref{eq:quant}) is not differentiable, we employ the straight-through gradient estimator following \cite{bengio2013estimating}, to copy the gradients from the decoder's input to the encoder's output.

\PAR{Stage 2.} Next, we train the AR condition predictor and diffusion head jointly for 200 iterations while keeping the VQ-VAE motion encoder fixed. 
The AR condition predictor is a two-layer transformer-based decoder with four attention heads, while the diffusion head is a single-layer MLP sharing the same hidden size. 
For this phase, we adopt a scaled linear schedule \cite{ho2020denoising} with a total of $\noisestep = 1000$ steps, 
and we employ the Denoising Diffusion Implicit Models (DDIM) \cite{song2020ddim}, using only 50 steps to sample during inference.

To better improve the quality of latents, we introduce a new latent loss function inspired by MAR \cite{li2024mar} that calculates the L1 distance between the generated latents $\tilde{z}$ and the ground truth latents $\zq$:
\begin{equation}
  \mathcal{L}_{latent} = \| \tilde{\zprm}_0^{\durationl} - \zprm_q^{\durationl} \|_1.
  \label{eq:diffloss}
\end{equation}
We then apply the following geometric losses in the 3D space: the vertex loss $\mathcal{L}_{vert}$ \cite{cudeiro2019voca} for the positions of the mesh vertices and the velocity loss $\mathcal{L}_{vel}$ \cite{cudeiro2019voca} for better temporal consistency:
\begin{equation}
  \mathcal{L}_{vert} = \| \tilde{\xbold}^{\durationl} - \xbold^{\durationl} \|_2^2,
  \label{eq:vertloss}
\end{equation}
\begin{equation}
  \mathcal{L}_{vel} = \| (\tilde{\xbold}^{\duraplus}-\tilde{\xbold}^{\duraminus}) - (\xbold^{\duraplus}-\xbold^{\duraminus}) \|_2^2.
\end{equation}

In summary, we combine all the aforementioned losses and apply a weighted sum to obtain the final loss function for stage 2 as:
\begin{equation}
  \mathcal{L}_{stage2} = \lambda_{latent}\mathcal{L}_{latent} + \lambda_{vert}\mathcal{L}_{vert} + \lambda_{vel}\mathcal{L}_{vel},
  \label{eq:stage2loss}
\end{equation} where $\lambda_{latent} = \lambda_{vert} = \lambda_{vel} = 1.0$. All models are trained using the AdamW \cite{loshchilov2017adamw} optimizer with a batch size of 1 and a learning rate of 0.0001 on a single Nvidia RTX 4090 GPU. More details of the network architecture are provided in the Appendix.
\section{Experiments and Results}

\subsection{Datasets}\label{sec:datasets}
We use two publicly available datasets, BIWI \cite{fanelli2010biwi} and VOCASET \cite{cudeiro2019voca},  to evaluate the effectiveness our method. 
Both datasets provide 3D scan-audio pairs of utterances spoken in English.

\PAR{BIWI Dataset.} The BIWI dataset is a comprehensive collection of affective speech and corresponding detailed 3D facial geometries. 
It consists of 14 participants who read 40 English sentences, each recorded in neutral and emotional contexts. 
The 3D facial data is captured at a rate of 25 fps, with each frame containing 23,370 vertices. 
The average duration of each sequence is approximately 4.67 seconds.

\PAR{VOCASET Dataset.} The VOCASET dataset includes 480 paired audio-visual sequences recorded from 12 subjects. 
The facial motion is captured at 60 frames per second and typically lasts about 4 seconds. 
Unlike the BIWI dataset, each 3D face mesh in VOCASET is registered to the FLAME \cite{li2017flame} topology, which consists of 5023 vertices. 

\subsection{Evaluation Metrics}\label{sec:metrics} 

Following previous works \cite{fan2022faceformer,xing2023codetalker}, we adopt two metrics for quantitative evaluation.
\PAR{Lip Vertex Error (LVE).} We measure lip synchronization following \cite{richard2021meshtalk} by calculating the L2 error between the predicted and ground-truth lip vertices.
\PAR{Face Dynamics Distance (FDD).} We follow \cite{xing2023codetalker} evaluating the deviation of upper face motions by calculating each upper face vertex's standard deviation of the element-wise L2 norm along the temporal axis. Smaller FDD indicates that the predicted expressions exhibit high consistency with the natural trends of facial dynamics. 
\PAR{Mouth Open Difference (MOD).} Following \cite{sun2024diffposetalk}, we compute the average absolute difference in mouth opening between the predicted and ground truth. A smaller value indicates better alignment in the degree of mouth opening. This helps reflect whether the model generates over-smooth mouth movements.

\begin{figure*}[t]
    \centering
    \includegraphics[trim=0cm 4cm 0cm 0cm,width=\linewidth]{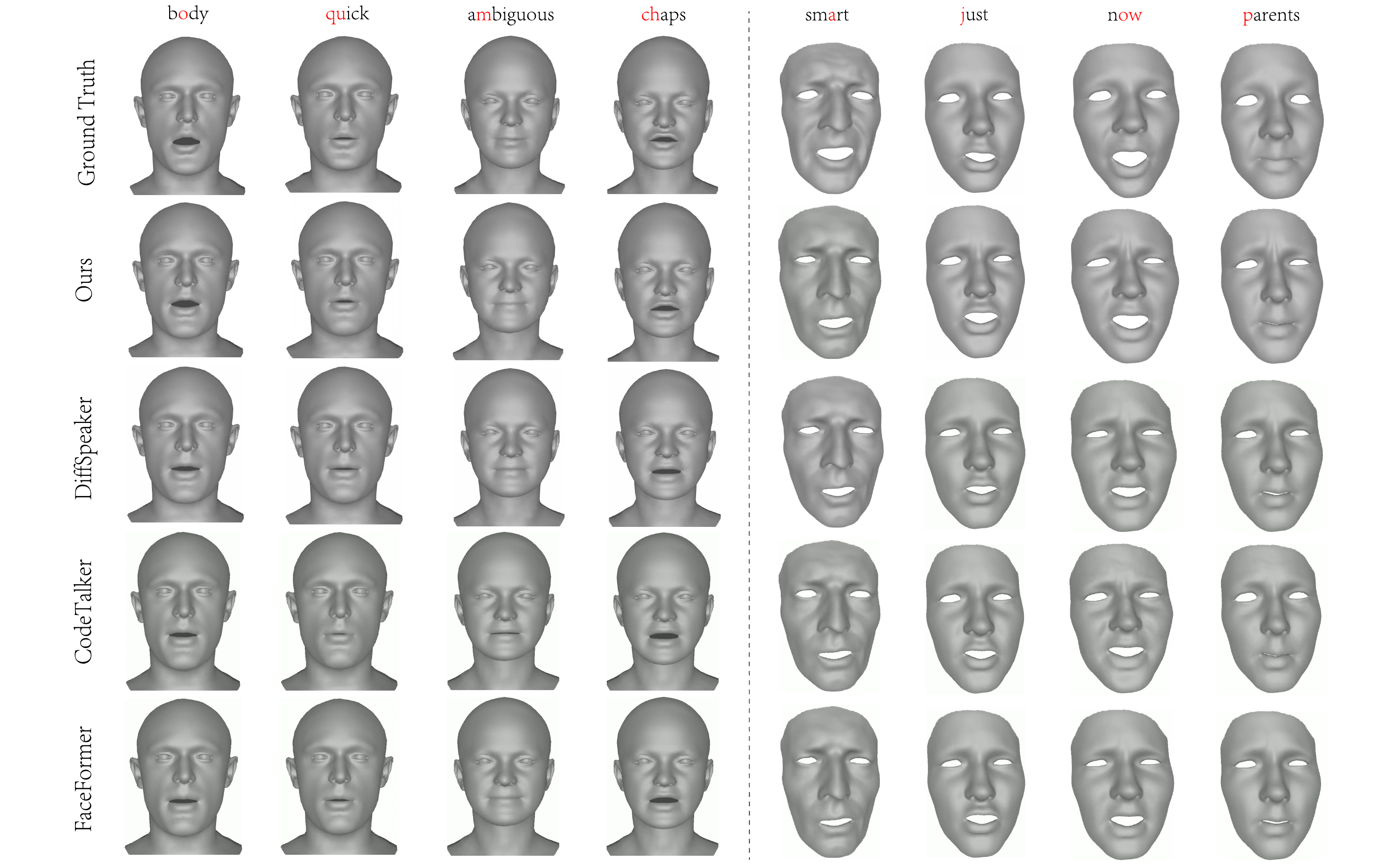}
    \vspace{+22pt}
    \caption{\textbf{Qualitative comparison with the state-of-the-arts.} The left side shows results on the VOCASET-Test dataset, while the right side shows results on the BIWI-Test-B dataset. \textcolor{red}{Red} words indicate phonemes being pronounced. Compared to other methods, our approach produces more natural lip shapes, with rounder mouth formations when pronouncing vowels like 'a', 'o', and 'u', and better lip closure for bilabial consonants such as 'm' and 'p'.}
    \label{fig:quality}
  \end{figure*}

\subsection{Quantitative Evaluation}\label{sec:quantitative}

\PAR{Baselines.} We compare our method with the following state-of-the-art methods: VOCA \cite{cudeiro2019voca}, MeshTalk \cite{richard2021meshtalk}, FaceFormer \cite{fan2022faceformer}, CodeTalker \cite{ng2022learning2listen}, FaceDiffuser \cite{stan2023facediffuser}, DiffSpeaker \cite{ma2024diffspeaker}, Imitator \cite{thambiraja2023imitator} and DiffPoseTalk \cite{sun2024diffposetalk}.
Among these, VOCA uses a MLP as decoder while MeshTalk uses a U-Net, both of which use convolutional neural networks to extract features from audio.
FaceFormer, CodeTalker and Imitator use transformers to model the long-term context.
FaceDiffuser uses diffusion to generate facial animation and applies GRU as the denoiser.  
DiffSpeaker chooses transformer decoder as the backbone network and outperforms previous methods.
DiffPoseTalk predicts FLAME parameters directly. To enable comparison under our vertex-based setting, we adapt its official implementation to output mesh vertices.
For all methods, we follow their original training and testing protocols as closely as possible, and apply official pre-trained weights if available.
For speakers unseen during training (i.e., not learned in the style embedding space), we generate multiple animations by conditioning on each speaker identity in training set separately, then compute evaluation metrics for each and report the average as the final result.

\begin{table}[t]
    \centering
    \caption{\textbf{Quantitative comparison} of StreamingTalker with baseline methods on two benchmark datasets. For VOCASET, LVE values are in units of $10^{-5}$mm. For BIWI, LVE values are in units of $10^{-4}$mm, while FDD values are in units of $10^{-5}$mm. MOD values are in units of $10^{-3}$mm for both datasets. Among them, \textbf{bold} indicates the best performance. $\downarrow$ means lower is better.}
    \resizebox{\linewidth}{!}{
    \begin{tabular}{lcccccc}
        \toprule
        \multirow{3}{*}{Methods} & \multicolumn{2}{c}{VOCASET} & \multicolumn{3}{c}{BIWI} \\
        \cmidrule(lr){2-3} \cmidrule(lr){4-6}
        & LVE $\downarrow$ & MOD $\downarrow$  & LVE $\downarrow$  & FDD $\downarrow$ & MOD $\downarrow$ \\
        \midrule
        VOCA & 4.9245 & 4.7131 & 6.5563 & 8.1816 & 33.7083 \\
        MeshTalk & 4.5441 & 4.4520 & 5.9181 & 5.1025 & 18.9012 \\
        FaceFormer & 4.1090 & 4.2001 & 5.3077 & 4.6408 &  16.7510 \\
        CodeTalker & 3.9445 & 4.1834 & 4.7914 & 4.1170 & 13.4410 \\
        FaceDiffuser & 4.1089 & 3.8069 & 4.2986 & 3.9098 & 8.6695 \\
        DiffSpeaker & 3.1478 & 3.5339 & 4.2829 & 3.8535 & \textbf{8.4091} \\
        Imitator & 3.1352 & 3.6610 & 4.3515 & 3.8066 & 10.3524 \\
        DiffPoseTalk & 3.9110 & 3.8819 & 5.3128 & 3.8256 & 12.6930 \\
        \midrule
        Ours  & \textbf{2.7206} & \textbf{3.4987} & \textbf{4.2504} & \textbf{3.6690} & 8.5208 \\
        \bottomrule
    \end{tabular}
    }
    \label{tab:quant}
\end{table}

We apply all metrics to the BIWI dataset but only LVE and MOD for the VOCASET dataset based on its limited facial expression variation. 
As shown in \Tabref{tab:quant}, our model achieves the best results on both LVE and FDD, demonstrating its ability to produce accurate lip movements and natural upper-face dynamics synchronized with speech.
Although the MOD is slightly higher than that of DiffSpeaker on short sequences, the difference is marginal and remains within an acceptable perceptual range.

Additionally, we evaluate the model's ability to generate extended facial motion sequences far beyond the training horizon.
Specifically, we use BIWI sequences of 2000 frames (i.e., over 60 seconds) to test long-term generalization.
As shown in \Tabref{tab:longseq}, our model consistently outperforms previous methods across all metrics, including MOD, indicating better temporal stability and coherence over long durations.
This result highlights the robustness of our autoregressive framework in modeling long-term dependencies without sacrificing motion quality or expressiveness.

\begin{table}[t]
    \centering
    \caption{\textbf{Comparison of results on long sequences} concatenated from BIWI-Test-B. The generated facial motion lengths are set to 2000 frames. The units used here are same as in \Tabref{tab:quant}.}
    \begin{tabular}{lccc}
        \toprule
        \multirow{1}{*}{Methods} & LVE $\downarrow$  & FDD $\downarrow$ & MOD $\downarrow$ \\
        \midrule
        FaceFormer & 5.4079 & 5.1418 & 20.7980 \\
        CodeTalker & 5.7064 & 6.6470 & 15.1021 \\
        FaceDiffuser & 5.3125 & 4.2674 & 9.0024 \\
        DiffSpeaker & 5.2213 & 4.7980 & 8.9528 \\
        \midrule
        Ours  & \textbf{4.4596} & \textbf{3.8912} & \textbf{8.8017} \\
        \bottomrule
    \end{tabular}
    \label{tab:longseq}
\end{table}

\subsection{Qualitative Evaluation}\label{sec:qualitative}

In the qualitative evaluation, we visually compare our method with FaceFormer, CodeTalker and DiffSpeaker. 
As shown in \Figref{fig:quality}, the animation results of different methods reveal that our approach generates lip movements that are more accurate and better synchronized with the audio. 
The mouth opens and closes more naturally compared to our competitors. For example, our method excels in handling rounded vowels, as demonstrated in the word “body” and “now”. 
Besides, for words like “quick”, “chaps” and “just” the model accurately generates a pursed lip shape. 
Furthermore, our method ensures proper lip closure in words like "ambiguous" and "parents", accurately modeling the /m/ and /p/ sounds that require the lips to be fully closed.
For more visual comparisons, please refer to our supplementary video.

\setlength{\tabcolsep}{8pt}
\begin{table*}[htbp]
    \centering
    \caption{\textbf{Ablation study.}  We compare our method with six variants to validate our main design choices. Among them, \textbf{bold} indicates the best results. $\downarrow$ means lower is better. Mean values are reported.
    }
    \begin{tabular}{lccccc}
        \toprule
        \multirow{3}{*}{Methods} & \multicolumn{3}{c}{BIWI} & \multicolumn{2}{c}{VOCASET} \\
        \cmidrule(lr){2-4} \cmidrule(lr){5-6}
        & LVE $\downarrow$  & FDD $\downarrow$ & MOD $\downarrow$ & LVE $\downarrow$ & MOD $\downarrow$ \\
        & ($\times 10^{-4}\, \text{mm}$) & ($\times 10^{-5}\, \text{mm}$) & ($\times 10^{-3}\, \text{mm}$) & ($\times 10^{-5}\, \text{mm}$) & ($\times 10^{-3}\, \text{mm}$) \\
        \midrule
        w/o Diffusion Head & $4.4297$ & $5.3934$ & $12.4698$ & $3.5070$ & $3.6011$ \\
        w/o VQ-VAE & $8.4578$ & $9.7510$ & $9.2511$ & $4.9521$ & $9.7510$\\
        Use VAE as encoder & $4.2686$ & $3.7010$ & $8.8292$ & $2.7516$ & $5.2660$\\
        Use all history motions & $4.7598$ & $4.0141$ & $8.7296$ & $2.8168$ & $3.5637$\\
        w/o Cross-Attetion & $11.7510$ & $6.9653$ & $21.2168$ & $7.4660$ & $10.6860$\\
        w/o Self-Attetion & $4.3025$ & $4.1083$ & $15.5354$ & $3.0321$ & $3.7499$\\
        \midrule
        Ours  & $\textbf{4.2504 }$ & $\textbf{3.6690 }$ & $\textbf{8.5208}$ & $\textbf{2.7206 }$ & $\textbf{3.4987}$ \\
        \bottomrule
    \end{tabular}
    \label{tab:ablation}
\end{table*}
\setlength{\tabcolsep}{6pt}

\subsection{Ablation Study}\label{sec:ablation}

As shown in \Tabref{tab:ablation}, we compare our method with five main ablation settings:

\noindent (1) \textbf{w/o Diffusion Head.}
To validate the effectiveness of the diffusion process, we remove the diffusion head and directly predict facial motion from the condition using an MLP. 
Without the diffusion head, the model's ability to capture the complex distribution of facial motion is reduced, resulting in degraded performance across all metrics.

\noindent (2) \textbf{Effect of Motion Latent Encoding.}
To evaluate the impact of motion latent encoding, we first remove the VQ-VAE and directly denoise the motion. This leads to a significant drop in motion quality, highlighting the importance of latent representation.
We further replace the VQ-VAE with a standard VAE to assess the model’s robustness to the choice of encoder. The results show only marginal performance differences, suggesting that our model remains stable even with continuous latent representations.

\noindent (3) \textbf{Use All History Motions.}
To evaluate the effectiveness of our fixed-length history strategy, we instead use all previous frames as historical context. This leads to slightly worse performance. We attribute this to a mismatch between training and testing conditions: the BIWI training sequences are relatively short, while test sequences are much longer. Using all history at test time exposes the model to significantly longer temporal dependencies than seen during training, causing a distribution shift that harms performance. 
In contrast, our fixed-window approach enforces a consistent context length, leading to better generalization across varying sequence durations.

\noindent (4) \textbf{w/o Cross-Attention.}
To evaluate the role of cross-attention layers, we remove all of them, eliminating the interaction between audio and motion features. 
Without audio guidance, the generated facial meshes become nonsensical and lack semantic consistency.

\noindent (5) \textbf{w/o Self-Attention.} 
To examine the impact of self-attention, we remove all self-attention layers. 
This results in a noticeable decline in global consistency, particularly in the upper face, where movements lose coherence, leading to lower FDD scores.

Additional qualitative results for the ablation study can be found in our supplementary materials.

\begin{figure}[htbp]
    \centering
    \includegraphics[trim=0cm 4cm 0cm 0cm,width=\linewidth]{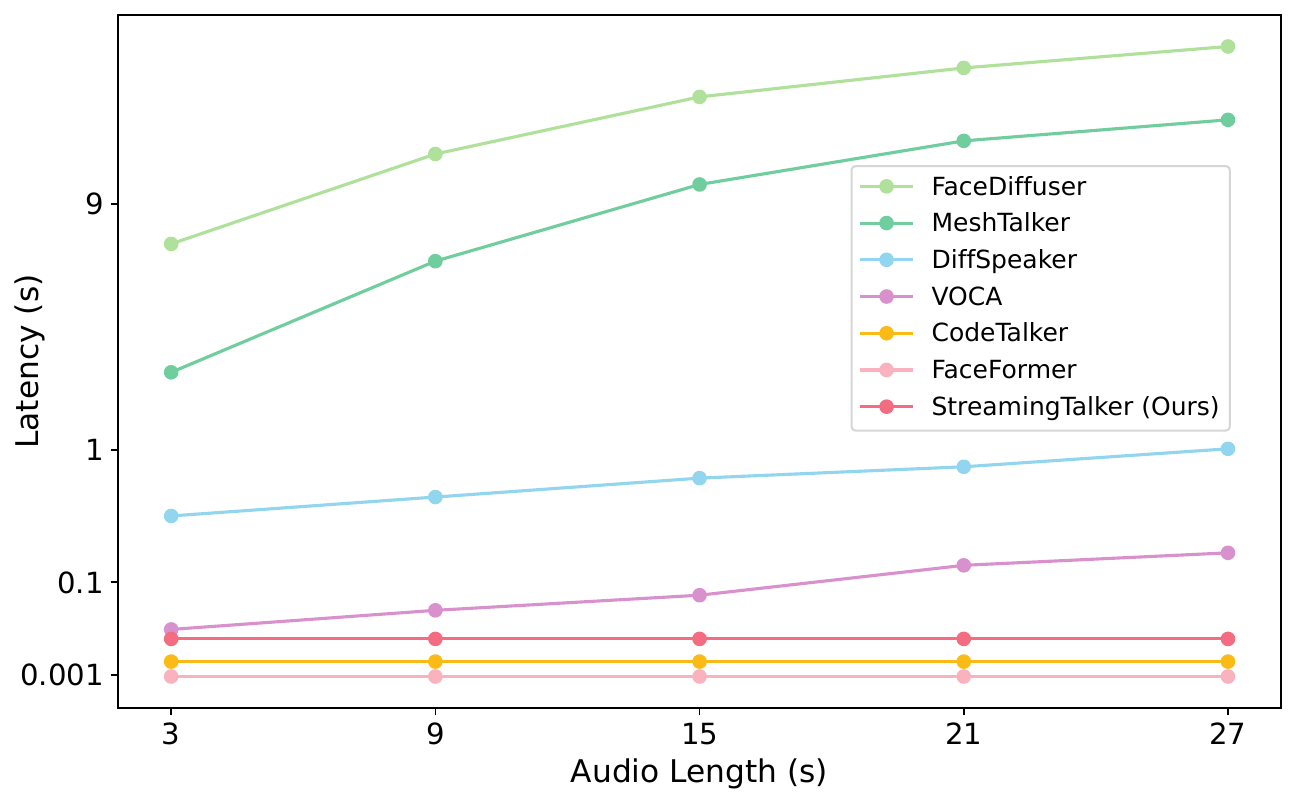}
    \vspace{+30pt}
    \caption{\textbf{Inference latency for 3-27 second audio clips.} The figure compares the performance of various models, including full sequence diffusion models (DiffSpeaker, FaceDiffuser), deterministic models (VOCA, MeshTalk), and AR models (FaceFormer, CodeTalker). Our model outperforms all non-AR models in terms of inference speed, maintaining consistent latency regardless of audio length.}
    \label{fig:inferencelatency}
\end{figure}

\subsection{Real-time Application}\label{sec:latency}
Inference latency is a critical factor in real-time applications, as it directly affects the user experience.
In this paper, we define inference latency as the time elapsed from the input of audio to the generation of the first frame of animation output.
We measured the inference latency for audio inputs of varying lengths on a 4090 GPU. 
The results, as shown in \Figref{fig:inferencelatency}, demonstrate that our method significantly outperforms all full sequence diffusion methods in terms of latency.
Additionally, our method maintains consistent inference latency across varying audio lengths.
This improvement is attributed to the unique design of our AR diffusion model, which supports streaming input. 
Unlike traditional diffusion methods that process the entire sequence at once, our model can begin rendering the output as soon as the first frame is processed. 
This process takes only 25 ms, thus meeting the requirements for real-time rendering.

We further build a real-time demo leveraging the streaming capability of our model. In this demo, a user speaks to ask a question, and the system first generates a response using a large language model (LLM). The response is then converted to speech via Google TTS \cite{gtts}, which is fed into our AR diffusion model to generate facial animation. Thanks to the model’s low-latency design, audio is processed in a streaming manner, with animation frames generated and rendered simultaneously at up to 40 FPS. 

Further details on per-component latency, implementation of our real-time demo, and the video demonstration can be found in the supplementary material.
\section{Discussions}
Although our proposed approach achieves significant improvements in generating 3D facial animations from speech, it still has some limitations.
Firstly, the current dataset consists of relatively short sequences, which are not sufficient to capture the full range of facial expressions a person can exhibit. 
As a result, our model is unable to extract long-term contextual information, so a larger and more diverse dataset could potentially improve the model's performance. 
Secondly, our model currently lacks emotional information, which could be a valuable addition to future research. 
Incorporating emotional cues, such as EMOCA \cite{danvevcek2022emoca}, could enhance the model's ability to generate more expressive animations and further improve the performance.

\section{Conclusion}
In this work, we introduce an AR diffusion model for speech-driven 3D facial animation. Our model dynamically integrates historical motion data, enhancing adaptability and context awareness compared to full sequence diffusion methods. This approach mitigates performance drops when generating sequences beyond the training horizon. Quantitative results show our method outperforms state-of-the-art techniques. We also developed an interactive real-time demo integrating an LLM for customizable attribute control and supporting streaming input, ensuring smooth, real-time rendering. Extensive ablation studies validate our network architecture and design choices.

\section{Acknowledgments}
This work was partially supported by the following grants: National Key R \& D Program of China (No. 2024YFB2809102), Zhejiang Provincial Natural Science Foundation of China (No. LR25F020003), Information Technology Center and State Key Lab of CAD \& CG, Zhejiang University.

\bibliography{aaai2026}

@inproceedings{rombach2022,
  title={High-resolution image synthesis with latent diffusion models},
  author={Rombach, Robin and Blattmann, Andreas and Lorenz, Dominik and Esser, Patrick and Ommer, Bj{\"o}rn},
  booktitle={Proceedings of the IEEE/CVF conference on computer vision and pattern recognition},
  pages={10684--10695},
  year={2022}
}

@inproceedings{xing2023codetalker,
  title={Codetalker: Speech-driven 3d facial animation with discrete motion prior},
  author={Xing, Jinbo and Xia, Menghan and Zhang, Yuechen and Cun, Xiaodong and Wang, Jue and Wong, Tien-Tsin},
  booktitle={Proceedings of the IEEE/CVF Conference on Computer Vision and Pattern Recognition},
  pages={12780--12790},
  year={2023}
}

@article{baevski2020wav2vec,
  title={wav2vec 2.0: A framework for self-supervised learning of speech representations},
  author={Baevski, Alexei and Zhou, Yuhao and Mohamed, Abdelrahman and Auli, Michael},
  journal={Advances in neural information processing systems},
  volume={33},
  pages={12449--12460},
  year={2020}
}

@article{hsu2021hubert,
  title={Hubert: Self-supervised speech representation learning by masked prediction of hidden units},
  author={Hsu, Wei-Ning and Bolte, Benjamin and Tsai, Yao-Hung Hubert and Lakhotia, Kushal and Salakhutdinov, Ruslan and Mohamed, Abdelrahman},
  journal={IEEE/ACM transactions on audio, speech, and language processing},
  volume={29},
  pages={3451--3460},
  year={2021},
  publisher={IEEE}
}

@article{bengio2013estimating,
  title={Estimating or propagating gradients through stochastic neurons for conditional computation},
  author={Bengio, Yoshua and L{\'e}onard, Nicholas and Courville, Aaron},
  journal={arXiv preprint arXiv:1308.3432},
  year={2013}
}

@inproceedings{panayotov2015librispeech,
  title={Librispeech: an asr corpus based on public domain audio books},
  author={Panayotov, Vassil and Chen, Guoguo and Povey, Daniel and Khudanpur, Sanjeev},
  booktitle={2015 IEEE international conference on acoustics, speech and signal processing (ICASSP)},
  pages={5206--5210},
  year={2015},
  organization={IEEE}
}

@article{li2024mar,
  title={Autoregressive Image Generation without Vector Quantization},
  author={Li, Tianhong and Tian, Yonglong and Li, He and Deng, Mingyang and He, Kaiming},
  journal={arXiv preprint arXiv:2406.11838},
  year={2024}
}

@article{ma2024diffspeaker,
  title={DiffSpeaker: Speech-Driven 3D Facial Animation with Diffusion Transformer},
  author={Ma, Zhiyuan and Zhu, Xiangyu and Qi, Guojun and Qian, Chen and Zhang, Zhaoxiang and Lei, Zhen},
  journal={arXiv preprint arXiv:2402.05712},
  year={2024}
}

@article{press2021train,
  title={Train short, test long: Attention with linear biases enables input length extrapolation},
  author={Press, Ofir and Smith, Noah A and Lewis, Mike},
  journal={arXiv preprint arXiv:2108.12409},
  year={2021}
}

@article{ho2020denoising,
  title={Denoising diffusion probabilistic models},
  author={Ho, Jonathan and Jain, Ajay and Abbeel, Pieter},
  journal={Advances in neural information processing systems},
  volume={33},
  pages={6840--6851},
  year={2020}
}

@article{tevet2209mdm,
  title={Human motion diffusion model. arXiv 2022},
  author={Tevet, G and Raab, S and Gordon, B and Shafir, Y and Cohen-Or, D and Bermano, AH},
  journal={arXiv preprint arXiv:2209.14916},
  year={2022}
}

@inproceedings{tseng2023edge,
  title={Edge: Editable dance generation from music},
  author={Tseng, Jonathan and Castellon, Rodrigo and Liu, Karen},
  booktitle={Proceedings of the IEEE/CVF Conference on Computer Vision and Pattern Recognition},
  pages={448--458},
  year={2023}
}

@inproceedings{stan2023facediffuser,
  title={Facediffuser: Speech-driven 3d facial animation synthesis using diffusion},
  author={Stan, Stefan and Haque, Kazi Injamamul and Yumak, Zerrin},
  booktitle={Proceedings of the 16th ACM SIGGRAPH Conference on Motion, Interaction and Games},
  pages={1--11},
  year={2023}
}

@article{massaro1212,
  title={12 Animated speech: research progress and applications},
  author={Massaro, DW and Cohen, MM and Tabain, M and Beskow, J and Clark, R},
  year={12}
}

@article{edwards2016jali,
  title={Jali: an animator-centric viseme model for expressive lip synchronization},
  author={Edwards, Pif and Landreth, Chris and Fiume, Eugene and Singh, Karan},
  journal={ACM Transactions on graphics (TOG)},
  volume={35},
  number={4},
  pages={1--11},
  year={2016},
  publisher={ACM New York, NY, USA}
}

@article{taylor2017deep,
  title={A deep learning approach for generalized speech animation},
  author={Taylor, Sarah and Kim, Taehwan and Yue, Yisong and Mahler, Moshe and Krahe, James and Rodriguez, Anastasio Garcia and Hodgins, Jessica and Matthews, Iain},
  journal={ACM Transactions On Graphics (TOG)},
  volume={36},
  number={4},
  pages={1--11},
  year={2017},
  publisher={ACM New York, NY, USA}
}

@article{karras2017audio,
  title={Audio-driven facial animation by joint end-to-end learning of pose and emotion},
  author={Karras, Tero and Aila, Timo and Laine, Samuli and Herva, Antti and Lehtinen, Jaakko},
  journal={ACM Transactions on Graphics (ToG)},
  volume={36},
  number={4},
  pages={1--12},
  year={2017},
  publisher={ACM New York, NY, USA}
}

@inproceedings{cudeiro2019voca,
  title={Capture, learning, and synthesis of 3D speaking styles},
  author={Cudeiro, Daniel and Bolkart, Timo and Laidlaw, Cassidy and Ranjan, Anurag and Black, Michael J},
  booktitle={Proceedings of the IEEE/CVF conference on computer vision and pattern recognition},
  pages={10101--10111},
  year={2019}
}

@inproceedings{richard2021meshtalk,
  title={Meshtalk: 3d face animation from speech using cross-modality disentanglement},
  author={Richard, Alexander and Zollh{\"o}fer, Michael and Wen, Yandong and De la Torre, Fernando and Sheikh, Yaser},
  booktitle={Proceedings of the IEEE/CVF International Conference on Computer Vision},
  pages={1173--1182},
  year={2021}
}

@inproceedings{fan2022faceformer,
  title={Faceformer: Speech-driven 3d facial animation with transformers},
  author={Fan, Yingruo and Lin, Zhaojiang and Saito, Jun and Wang, Wenping and Komura, Taku},
  booktitle={Proceedings of the IEEE/CVF Conference on Computer Vision and Pattern Recognition},
  pages={18770--18780},
  year={2022}
}

@article{vaswani2017attention,
  title={Attention is all you need},
  author={Vaswani, A},
  journal={Advances in Neural Information Processing Systems},
  year={2017}
}

@article{van2017vqvae,
  title={Neural discrete representation learning},
  author={Van Den Oord, Aaron and Vinyals, Oriol and others},
  journal={Advances in neural information processing systems},
  volume={30},
  year={2017}
}

@inproceedings{ng2022learning2listen,
  title={Learning to listen: Modeling non-deterministic dyadic facial motion},
  author={Ng, Evonne and Joo, Hanbyul and Hu, Liwen and Li, Hao and Darrell, Trevor and Kanazawa, Angjoo and Ginosar, Shiry},
  booktitle={Proceedings of the IEEE/CVF Conference on Computer Vision and Pattern Recognition},
  pages={20395--20405},
  year={2022}
}

@inproceedings{sohl2015diffusion,
  title={Deep unsupervised learning using nonequilibrium thermodynamics},
  author={Sohl-Dickstein, Jascha and Weiss, Eric and Maheswaranathan, Niru and Ganguli, Surya},
  booktitle={International conference on machine learning},
  pages={2256--2265},
  year={2015},
  organization={PMLR}
}

@article{goodfellow2020gan,
  title={Generative adversarial networks},
  author={Goodfellow, Ian and Pouget-Abadie, Jean and Mirza, Mehdi and Xu, Bing and Warde-Farley, David and Ozair, Sherjil and Courville, Aaron and Bengio, Yoshua},
  journal={Communications of the ACM},
  volume={63},
  number={11},
  pages={139--144},
  year={2020},
  publisher={ACM New York, NY, USA}
}

@article{kingma2013vae,
  title={Auto-encoding variational bayes},
  author={Kingma, Diederik P},
  journal={arXiv preprint arXiv:1312.6114},
  year={2013}
}

@article{sun2024diffposetalk,
  title={Diffposetalk: Speech-driven stylistic 3d facial animation and head pose generation via diffusion models},
  author={Sun, Zhiyao and Lv, Tian and Ye, Sheng and Lin, Matthieu and Sheng, Jenny and Wen, Yu-Hui and Yu, Minjing and Liu, Yong-jin},
  journal={ACM Transactions on Graphics (TOG)},
  volume={43},
  number={4},
  pages={1--9},
  year={2024},
  publisher={ACM New York, NY, USA}
}

@article{lin2024glditalker,
  title={GLDiTalker: Speech-Driven 3D Facial Animation with Graph Latent Diffusion Transformer},
  author={Lin, Yihong and Fan, Zhaoxin and Xiong, Lingyu and Peng, Liang and Li, Xiandong and Kang, Wenxiong and Wu, Xianjia and Lei, Songju and Xu, Huang},
  journal={arXiv preprint arXiv:2408.01826},
  year={2024}
}

@inproceedings{du2023dae,
  title={Dae-talker: High fidelity speech-driven talking face generation with diffusion autoencoder},
  author={Du, Chenpeng and Chen, Qi and He, Tianyu and Tan, Xu and Chen, Xie and Yu, Kai and Zhao, Sheng and Bian, Jiang},
  booktitle={Proceedings of the 31st ACM International Conference on Multimedia},
  pages={4281--4289},
  year={2023}
}

@article{fanelli2010biwi,
  title={A 3-d audio-visual corpus of affective communication},
  author={Fanelli, Gabriele and Gall, Juergen and Romsdorfer, Harald and Weise, Thibaut and Van Gool, Luc},
  journal={IEEE Transactions on Multimedia},
  volume={12},
  number={6},
  pages={591--598},
  year={2010},
  publisher={IEEE}
}

@article{li2017flame,
  title={Learning a model of facial shape and expression from 4D scans.},
  author={Li, Tianye and Bolkart, Timo and Black, Michael J and Li, Hao and Romero, Javier},
  journal={ACM Trans. Graph.},
  volume={36},
  number={6},
  pages={194--1},
  year={2017}
}

@article{loshchilov2017adamw,
  title={Fixing weight decay regularization in adam},
  author={Loshchilov, Ilya and Hutter, Frank and others},
  journal={arXiv preprint arXiv:1711.05101},
  volume={5},
  year={2017}
}

@article{song2020ddim,
  title={Denoising diffusion implicit models},
  author={Song, Jiaming and Meng, Chenlin and Ermon, Stefano},
  journal={arXiv preprint arXiv:2010.02502},
  year={2020}
}

@article{zhang2018mode,
  title={Mode-adaptive neural networks for quadruped motion control},
  author={Zhang, He and Starke, Sebastian and Komura, Taku and Saito, Jun},
  journal={ACM Transactions on Graphics (TOG)},
  volume={37},
  number={4},
  pages={1--11},
  year={2018},
  publisher={ACM New York, NY, USA}
}

@article{razavi2019vqvae2,
  title={Generating diverse high-fidelity images with vq-vae-2},
  author={Razavi, Ali and Van den Oord, Aaron and Vinyals, Oriol},
  journal={Advances in neural information processing systems},
  volume={32},
  year={2019}
}

@inproceedings{danvevcek2022emoca,
  title={Emoca: Emotion driven monocular face capture and animation},
  author={Dan{\v{e}}{\v{c}}ek, Radek and Black, Michael J and Bolkart, Timo},
  booktitle={Proceedings of the IEEE/CVF Conference on Computer Vision and Pattern Recognition},
  pages={20311--20322},
  year={2022}
}

@inproceedings{ramesh2021zero,
  title={Zero-shot text-to-image generation},
  author={Ramesh, Aditya and Pavlov, Mikhail and Goh, Gabriel and Gray, Scott and Voss, Chelsea and Radford, Alec and Chen, Mark and Sutskever, Ilya},
  booktitle={International conference on machine learning},
  pages={8821--8831},
  year={2021},
  organization={Pmlr}
}

@book{socket,
  title={Computer Networking: A Top-Down Approach},
  author={James Kurose and Keith Ross},
  year={2017},
  publisher={Pearson}
}

@misc{gtts,
  author={pndurette},
  title={gTTS: Google Text-to-Speech},
  year={2021},
  url={https://github.com/pndurette/gTTS},
  note={Accessed: 2025-01-23}
}

@misc{vosk_api,
  author = {Shmyrev, N. V.},
  title = {Vosk speech recognition toolkit: offline speech recognition API for android, iOS, Raspberry Pi and servers with Python, Java, C\# and Node},
  year = {2023},
  url = {https://github.com/alphacep/vosk-api},
  urldate = {2024-01-23}
}

@inproceedings{thambiraja2023imitator,
  title={Imitator: Personalized speech-driven 3d facial animation},
  author={Thambiraja, Balamurugan and Habibie, Ikhsanul and Aliakbarian, Sadegh and Cosker, Darren and Theobalt, Christian and Thies, Justus},
  booktitle={Proceedings of the IEEE/CVF international conference on computer vision},
  pages={20621--20631},
  year={2023}
}

\section{Ethics Statements}
As our approach enables realistic synthesis of facial motion from audio, we acknowledge potential risks of misuse, such as in generating deceptive or unauthorized content.
We emphasize that our work is intended for legitimate applications, including virtual communication, accessibility tools, and creative production.



\clearpage
\setcounter{page}{1}
\appendix

\section{Appendix}
\subsection{A. Dataset Details}
We further elaborate on the details of the datasets, including the dataset splits and the size of the datasets.

\PAR{BIWI Dataset.} In our experiments, the dataset is divided into a training set (BIWI-Train) comprising 192 sentences spoken by six participants following \cite{fan2022faceformer}, with each participant contributing 32 sentences. 
The validation set (BIWI-Val) includes 24 sentences from the same six participants, with each speaking four sentences. 
Additionally, there are two testing sets: BIWI-Test-A, which contains 24 sentences from six participants seen during training, and BIWI-Test-B, which includes 32 sentences from eight participants not seen during training. 
BIWI-Test-A is suitable for quantitative and qualitative evaluations, while BIWI-Test-B is primarily used for qualitative assessment.
\PAR{VOCA Dataset.} For consistency and fair comparison, we adopt the same training (VOCA-Train), validation (VOCA-Val), and testing (VOCA-Test) splits as used in VOCA \cite{cudeiro2019voca} and \cite{fan2022faceformer}.


\subsection{B. Implementation Details}

\PAR{Network architecture.}
In our experiments on the VOCASET dataset, we use a six-layer transformer architecture with eight attention heads for both the encoder and decoder of the VQ-VAE. 
The feature dimension is set to 1024 and the feedforward network dimension is 1536, 
while the size of the codebook is 256, with the number of face components set to 16. 
For the AR Condition Predictor, we use a two-layer transformer decoder with four attention heads, whose hidden size is 768 and the feedforward network size is 1024 while the history length is set to 60. 
Furthermore, The diffusion head is implemented as a single-layer MLP with the same hidden size as the AR Condition Predictor. 
We adopt a scaled linear schedule \cite{ho2020denoising} for the diffusion process, with a total of $\noisestep = 1000$ steps. 
In the denoising phase, we employ the Denoising Diffusion Implicit Models (DDIM) \cite{song2020ddim}, using only 50 steps to achieve efficient denoising while maintaining high-quality results.

For experiments on the more complex BIWI dataset, which contains more intricate data with greater upper-face movements, we scale up the model. 
Specifically, we increase the size of the codebook to 512, and the number of face components is set to 8 to mitigate the risk of overfitting \cite{xing2023codetalker}. 
The AR Condition Predictor is also scaled accordingly, with a hidden size of 1024 and a feedforward network size of 2048 while the history length is set to 120. 
In addition, we fine-tune the VAE decoder to better capture the rich facial expressions present in the BIWI dataset. 
All other configurations remain consistent with those used for the VOCASET dataset.

\PAR{Training.}
Our model was developed using PyTorch and trained on a single NVIDIA RTX 4090 GPU. 
We set the batch size to 1 and train 400 epochs using the AdamW \cite{loshchilov2017adamw} optimizer with a learning rate initialized to 1e-4 in both stages. 
In the first stage of training, we decrease the learning rate by half every 20 epochs in VOCASET and 40 epochs in BIWI.
The overall training takes about 12 hours.

\subsection{C. Real-time Demo}

We have implemented a real-time interactive demo that consists of two main components: the client and the server, which can be easily deployed on a computer equipped with an NVIDIA RTX 3090 GPU or above.

\begin{figure}[t]
    \centering
    \includegraphics[trim=0cm 4cm 0cm 0cm,width=\linewidth]{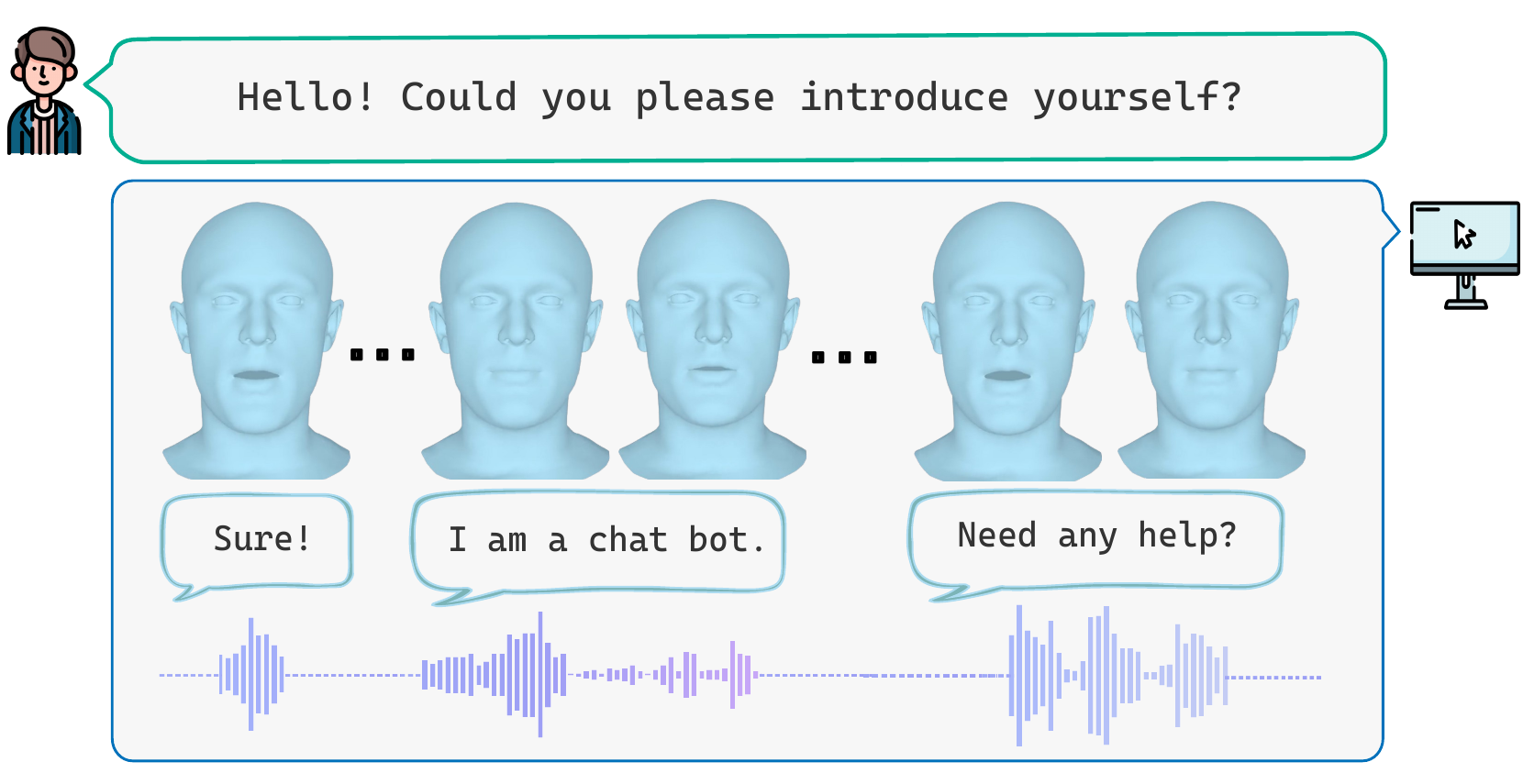}
    \vspace{+25pt}
    \caption{\textbf{Overview of our real-time demo system.}}
    \label{fig:teaser}
\end{figure}

\PAR{Client}

The client is responsible for capturing the user's audio input and converting it to text. To achieve this, we use a local Vosk \cite{vosk_api} model, which performs speech-to-text (STT) processing. Once the audio is transcribed into text, the client sends the resulting text to the server via a Socket \cite{socket} connection for further processing.

\PAR{Server}

The server side of the demo is composed of two key components: our autoregressive AR diffusion model and a large language model (LLM). Upon receiving the text from the client, the server first utilizes the LLM to generate a relevant response based on the input text.
Next, the server converts this response into speech using the Google Text-to-Speech \cite{gtts} tool. This text-to-speech conversion allows the system to produce an audio response aligned with the generated text.

To optimize the real-time performance of the system, we employ multithreading during the model inference stage. In the main thread, the model processes the audio using the Wav2Vec2 \cite{baevski2020wav2vec} processor, which handles the pre-processing of the speech input. Following this, the AR diffusion model performs streaming generation, producing a sequence of frames that represent the facial animation corresponding to the audio.
Each generated frame is added to a global queue, allowing the frames to be processed in parallel. The render thread continuously retrieves frames from the queue and visualizes them in real time. This ensures the facial animation remains synchronized with the input speech and responds dynamically to the model's output.
To ensure the visual and audio components are properly synchronized, we implement a dedicated speech thread. This thread starts playing the audio once the first frame has been rendered and outputted by the render thread. By managing the audio playback this way, we maintain perfect synchronization between the animation and the corresponding speech, enhancing the user experience during the demo.

\subsection{D. Inference Time and Latency Analysis}

In this section, we clarify the inference time and latency of our system. 

\begin{table*}[t]
    \centering
    \caption{\textbf{Latency Breakdown of Different System Components.} 
    This table presents the time consumption of each component in our pipeline. 
    The total inference latency includes HuBERT feature extraction and the first-frame inference time, 
    while the inference time per frame considers only the components involved in generating each subsequent frame.}
    \label{tab:latency}
    \begin{tabular}{lcc}
        \toprule
        \textbf{Component} & \textbf{Time (ms)} & \textbf{Contribution} \\
        \midrule
        HuBERT & 5  & Part of inference latency \\
        VAE Encoder + Decoder & 5  & Part of per-frame inference time \\
        AR Condition Predictor & 3  & Part of per-frame inference time \\
        Diffusion Head & 12 & Part of per-frame inference time \\
        \midrule
        \textbf{Total inference latency} & \textbf{25} & (HuBERT + First-frame inference) \\
        \textbf{Total inference time per frame} & \textbf{20} & (VAE + AR Condition Predictor + Diffusion Head) \\
        \bottomrule
    \end{tabular}
\end{table*}

\PAR{Inference Latency}
Inference latency is defined as the time elapsed from when the model receives an input to when the first frame is generated. In our system, the total inference latency is 25 ms, which includes 5 ms for HuBERT feature extraction and 20 ms for model inference.

\PAR{Inference Time Per Frame}
Inference time per frame refers to the time that out model requires to generate each subsequent frame. In our system, the inference time per frame is 20 ms. This consists of 5 ms for the VAE encoder and decoder, 3 ms for the autoregressive condition predictor and 12 ms for the diffusion head. 

Table~\ref{tab:latency} provides a detailed breakdown of the latency contributions from different system components.

\subsection{E. Ablation Study}
\begin{figure*}[t]
    \centering
    \includegraphics[trim=0cm 4cm 0cm 0cm,width=\linewidth]{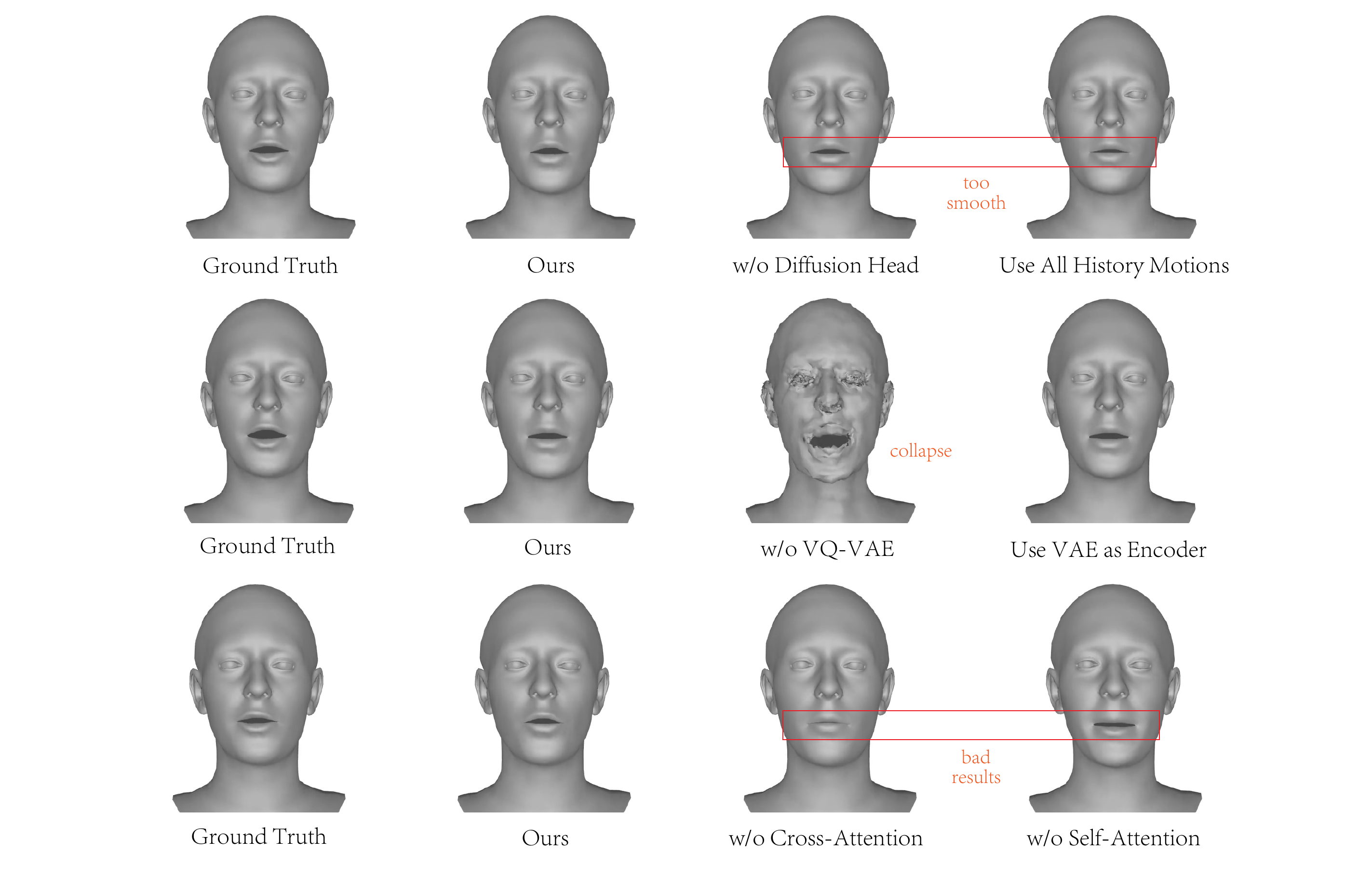}
    \vspace{+25pt}
    \caption{\textbf{Ablation study on key components of the model.} We perform six ablation experiments to evaluate the impact of different components. 1) \textbf{w/o Diffusion Head:} Removing the diffusion head significantly degrades model performance. 2) \textbf{Use All History Motions:} Omitting this results in overly smooth outputs. 3) \textbf{w/o VQ-VAE:} Without VQ-VAE, the model tends to collapse. 4) \textbf{Use VAE as Encoder:} No significant difference is observed, indicating our model is robust to the choice of encoder. 5) \textbf{w/o Cross-Attention:} Removing cross attention leads to erroneous results, as it is essential for incorporating speech information. 6) \textbf{w/o Self-attention:} Without self-attention, the model fails to maintain consistency.}
    \label{fig:ablation}
\end{figure*}

We conduct a comprehensive ablation study to evaluate the contribution of each core component in our framework. First, removing the diffusion head (\textbf{w/o Diffusion Head}) leads to a substantial performance drop, demonstrating its crucial role in refining motion details and ensuring stable generation quality. Second, using the entire history motion instead of a fixed-length window (\textbf{Use All History Motions}) produces overly smoothed facial motions, indicating that an appropriately selected history window is necessary to preserve dynamics while preventing over-conditioning. Third, removing the VQ-VAE module (\textbf{w/o VQ-VAE}) causes the model to collapse during training, underscoring its importance in stabilizing the latent space and providing a compact, discrete representation. Fourth, replacing VQ-VAE with a standard VAE encoder (\textbf{Use VAE as Encoder}) yields no significant difference in performance, suggesting that our overall framework is robust to the choice of encoding method. Fifth, eliminating the cross-attention mechanism (\textbf{w/o Cross-Attention}) severely damages the synchronization and expressiveness of the generated motions, as cross-attention is essential for integrating speech features into the generation process. Finally, removing self-attention (\textbf{w/o Self-Attention}) prevents the model from maintaining temporal consistency, leading to jittery or incoherent results.
The results of the ablation study are shown in Figure \ref{fig:ablation}.

\subsection{F. Qualitative Results}
\begin{figure*}[t]
    \centering
    \includegraphics[trim=0cm 4cm 0cm 0cm,width=\linewidth]{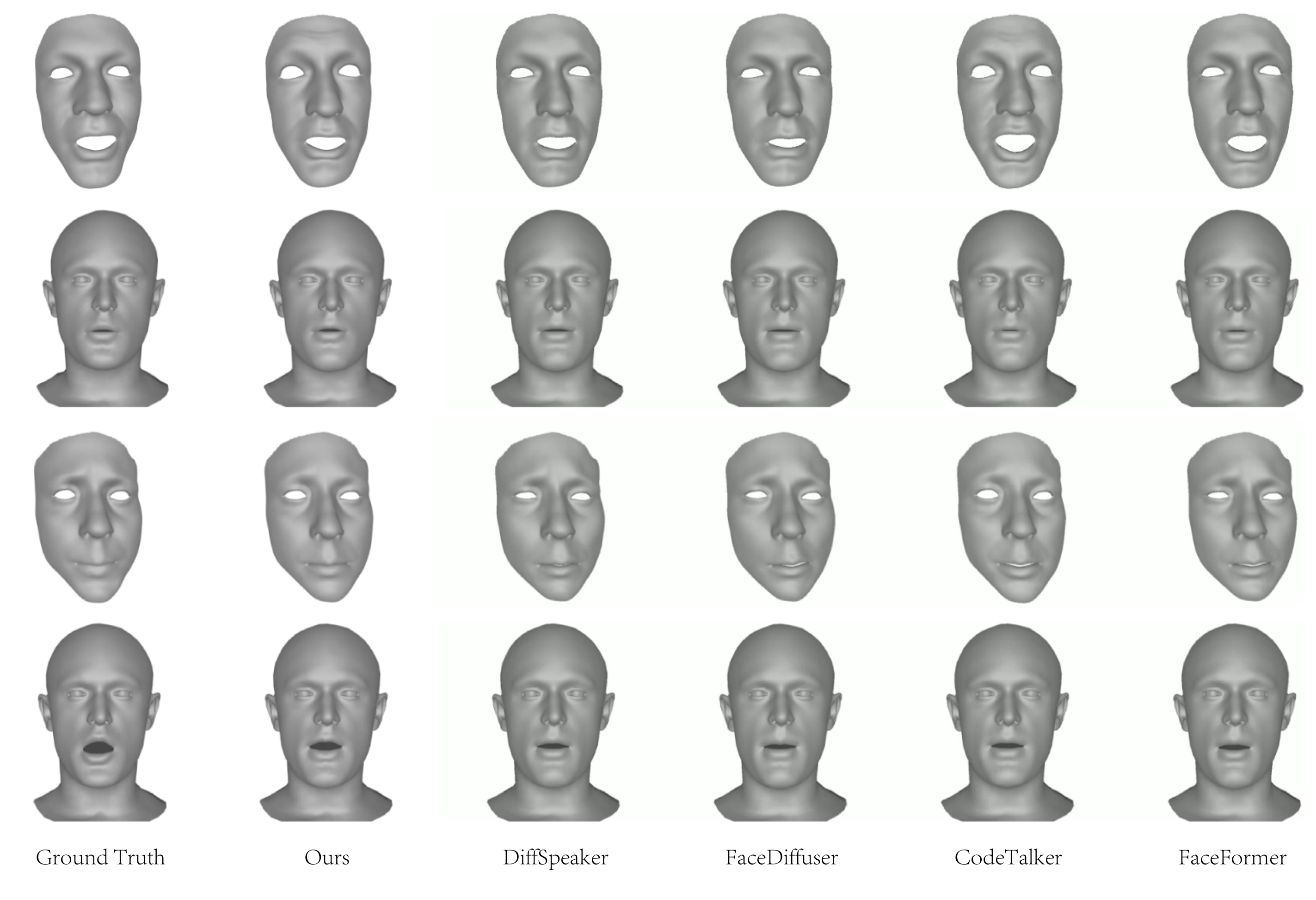}
    \vspace{+25pt}
    \caption{\textbf{Additional qualitative comparisons with previous state of the art methods.} Our method generates more natural and rounded mouth shapes for vowel-like sounds (e.g., “o,” “u”) and achieves accurate lip closure for bilabial consonants (e.g., “m,” “b,” “p”), showing clearer articulation than previous approaches. More visual results will be provided on our project page.}
    \label{fig:add}
\end{figure*}

We additionally provide qualitative comparisons with previous methods to visually assess the expressiveness and accuracy of the generated facial motions. As shown in the Figure \ref{fig:add}, our method produces noticeably more natural and rounded mouth shapes for vowel-like syllables such as “o” and “u,” while achieving accurate and complete lip closure for bilabial consonants including “m,” “b,” and “p.” These improvements highlight the effectiveness of our autoregressive diffusion framework in capturing fine-grained articulatory details. To further demonstrate the visual quality and temporal consistency of our results, we will prepare a video containing more comprehensive visualizations and upload it, with the access link provided on our GitHub page.

\subsection{G. Future Work}
Although our method achieves real-time and high-quality speech-driven facial animation, it still has several limitations. First, the identity embedding is tightly coupled with the individuals in the training set, making it difficult for the model to generalize to unseen identities without additional fine-tuning. Second, the current framework focuses primarily on lip motions and does not generate rich facial expressions, which limits the realism and expressiveness of the final animation. Future work may explore more flexible identity representations to enable better generalization, as well as incorporate style control or expression modeling to generate more expressive and personalized facial motions.

\end{document}